\documentclass[preprint,12pt,authoryear]{elsarticle}

\usepackage{natbib}
\usepackage{url}
\usepackage{cleveref}
\usepackage{subcaption}
\usepackage{booktabs}
\usepackage{makecell}
\usepackage{comment}
\usepackage{rotating}

\journal{Journal of Cleaner Production (forthcoming)}

\begin{document}
\let\WriteBookmarks\relax
\def\floatpagepagefraction{1}
\def\textpagefraction{.001}

\title{{Artificial Intelligence for Sustainability: Facilitating Sustainable Smart Product-Service Systems with Computer Vision}}

\author[1]{Jannis Walk}
\ead{jannis.walk@kit.edu}
\address[1]{Karlsruhe Institute of Technology, Kaiserstraße 89, 76133 Karlsruhe, Germany}
\address[2]{University of Bayreuth, Witelsbacherring 10, 95444 Bayreuth, Germany}
\address[3]{University of Illinois at Urbana-Champaign, Champaign, IL, United States}
\address[4]{Plansee SE, Metallwerk-Plansee-Straße 71, 6600 Reutte, Austria}

\author[2]{Niklas Kühl\textsuperscript{*}}
\ead{kuehl@uni-bayreuth.de}

\author[3]{Michael Saidani}
\ead{msaidani@illinois.edu}

\author[4]{Jürgen Schatte}
\ead{Juergen.Schatte@plansee.com}

\begin{abstract}
\renewcommand*{\thefootnote}{\fnsymbol{footnote}}
\footnote{corresponding author at: University of Bayreuth, Witelsbacherring 10, 95444 Bayreuth, Germany}
\renewcommand*{\thefootnote}{\arabic{footnote}}
Recent advances in artificial intelligence in general, and deep learning in particular, enable innovations that have a massive impact on society and industries. Autonomous driving, facial recognition, drug discovery, and speech recognition are examples of fundamental innovations facilitated by deep learning. However, the usage and impact of deep learning for cleaner production and sustainability purposes remain little explored. This work shows how deep learning can be harnessed to increase sustainability in production and product usage. Specifically, we utilize deep learning-based computer vision to determine the wear states of products. The resulting insights serve as a basis for novel product-service systems with improved integration and result orientation. Moreover, these insights are expected to facilitate product usage improvements and R\&D innovations. We demonstrate our approach on two products: machining tools and rotating X-ray anodes. From a technical standpoint, we show that it is possible to recognize the wear state of these products using deep-learning-based computer vision. In particular, we detect wear through microscopic images of the two products. We utilize a U-Net for semantic segmentation to detect wear based on pixel granularity. The resulting mean dice coefficients of 0.631 and 0.603 demonstrate the feasibility of the proposed approach. Consequently, experts can now make better decisions, for example, to improve the machining process parameters. To assess the impact of the proposed approach on environmental sustainability, we perform life cycle assessments that show gains for both products. The results indicate that the emissions of CO\textsubscript{2} equivalents are reduced by 12\% for machining tools and by 44\% for rotating anodes. This work can serve as a guideline and inspire researchers and practitioners to utilize computer vision in similar scenarios to develop sustainable smart product-service systems and enable cleaner production.
\end{abstract}


\maketitle

\section{Introduction}
In addition to the agricultural and energy sectors, the manufacturing industry is one of the largest emitters of CO\textsubscript{2} \citep{edenhofer2014summary} and its emissions are continuously rising \citep{hatfield2017assessing}. Therefore, there is an urgent need to reduce the carbon footprint of the manufacturing industry \citep{park2009energy}. An option to address this challenge is to substantially reduce waste generation through prevention, reduction, recycling, and reuse, as formulated by the United Nations \citep[p. 27]{Nations2015}. However, technical innovations enabling and supporting these activities are an imperative \citep{mohmmed2019driving}. 

In recent years, there have been major breakthroughs in the field of artificial intelligence (AI) systems based on deep learning (DL), which have enabled them to surpass human performance in specific tasks \citep{silver2017mastering, he2015delving}. As described by \citet{Vinuesa2020}, AI can positively affect sustainable development and multiple frameworks help to structure these endeavours \citep{ren2019comprehensive,zhang2017framework}. However, thus far, only a few real-world implementations of AI address sustainable development goals.
\enlargethispage*{1cm}
To address this research gap, this work applies DL-based computer vision (CV) to facilitate initiatives that address the aforementioned challenges. Specifically, we utilize CV to efficiently determine the wear states of products. As machine learning (ML) and DL has been shown to yield benefits in the related field of smart manufacturing \citep{Wang2018DLForSmartManufacturing, flath2018towards, migueis2022automatic}, it promises to be effective for the proposed approach. 

The products investigated in our case studies are machining tools used for machining processes and rotating anodes used in diagnostic imaging applications. These products show different types of wear over their life cycles, which can only be observed accurately with a microscope. We employ a DL-based CV model, which is a specialized AI technique, to determine the wear state of these products from microscopic images. 
For the associated case studies related to the products under consideration, complete automation is neither possible nor desirable. Instead, we rely on combining the strengths of human and artificial intelligence---an approach called hybrid intelligence \citep{Dellermann2019}. The DL-based CV model detects product wear from images in a reproducible and efficient manner. This task is tedious and difficult for humans to perform. However, final decisions, such as the adaptation of machining process parameters, are at the discretion of human experts. They can incorporate a plethora of additional information, such as operating conditions and years of domain expertise. At present, training an AI to incorporate this additional information is infeasible because a large amount of data is required to reflect all the nuances of real-world situations.

The results of the wear analysis performed through CV facilitate different types of product-service systems (PSSs) that support environmental sustainability in four ways. First, it can facilitate typical reversible strategies (4R): re-design, remanufacturing, reuse, and recycling \citep{Li2021}. 
Understanding the current wear state of a product is crucial for deciding which one of the 4Rs is most suitable from an economic and environmental sustainability perspective.
Second, the product usage stages can be improved in terms of environmental sustainability.
Data on the usage of products are scarce and generally of low quality because the products are typically owned by customers at this stage \citep{zhang2017framework}. The proposed approach provides information about this important phase, and thus, it enables more data-driven product lifecycle management. Third, assessing the wear state of products using CV can improve the usage of the same product in future iterations. For instance, \citet{Wang2018} show for steel that altering the usage stage is one of the biggest levers of manufacturing companies regarding sustainability. Finally, result-oriented PSSs can yield sustainability benefits \citep{Tukker2015}. In result-oriented PSSs, the client and provider agree on an outcome, but the provider can choose how the outcome will be achieved---in particular, no predefined product is involved \citep{tukker2004eight}. However, a detailed understanding of product usage is necessary to offer result-oriented PSSs because it is crucial to assess the risks and costs to make a competitive yet profitable offering \citep{tukker2004eight}. An accurate assessment of the wear state of products using CV can facilitate a detailed understanding of product usage.

Conceptually, the proposed approach depicts a novel type of sustainable smart product-service system as proposed by \cite{Li2021}. In contrast to previous studies by \citet{zhang2017framework} and \citet{Li2021}, the products under consideration do not have to be smart for the proposed approach; that is, no sensors are integrated in or placed on the product.

This work contributes to the existing literature by demonstrating the effectiveness of DL-based CV models for extracting valuable information from non-smart products to improve environmental sustainability. We validate this approach using two products by demonstrating its technical feasibility through an experiment and environmental sustainability through a life cycle assessment (LCA). The technical feasibility for detecting wear on the two products is demonstrated, and the environmental sustainability benefit of the proposed approach is verified through LCAs. Furthermore, the requirements of this approach are conceptualized to provide researchers and practitioners with guidance regarding its applicability to other PSSs.

The remainder of this work is organized as follows. In \Cref{sec:Materials_and_Methods}, we introduce relevant foundations and the methodology employed. Subsequently, in \Cref{sec:Results}, we present the evaluation and the results of the CV models and the LCAs. In \Cref{sec:Discussion}, we discuss our work in a broader context and conceptualize it. Finally, in \Cref{sec:Conclusion}, we summarize our work and discuss its limitations and possible future research.

\newpage
\section{Materials and Methods}
\label{sec:Materials_and_Methods}
This chapter introduces the relevant foundations in the fields of PSSs, CV, and DL. We then present the methodology and the selected case studies.
\subsection{Foundations}
\label{sec:Foundations}
In the following section, we first discuss sustainable PSSs and then introduce the fundamentals of CV and DL.
\subsubsection{Sustainable Smart Product-Service Systems}
PSSs were first defined in \citeyear{goedkoop1999product} by \citeauthor{goedkoop1999product} as ``a marketable set of products and services capable of jointly fulfilling a user's need.'' In the early stages of PSSs, sustainability was already an important concept \citep{li2020state}. In 2016, based on their literature review, \citet{Annarelli2016} define a PSS as ``a business model
focused toward the provision of a marketable set of products and services, designed to be economically, socially, and environmentally sustainable, with the final aim of fulfilling a customer's needs.'' More recently, various authors extended the concept of sustainable PSSs to be smart based on technological innovations, such as AI or Internet of things-based connectivity \citep{de2019overcoming,sakao2020product,Li2021}. \citet{Alcayaga2019} coined the term smart-circular systems, which they conceptualized as a combination of circular strategies, smart products, and PSSs. Based on \cite{EMF2016}, smart products are considered to possess the ability to sense, store, and communicate information about their environments and themselves. 
\citet{Li2021} propose a data-driven reversible framework for achieving sustainable smart PSSs. They illustrated this framework by sustainably developing a 3D printer.
This work contributes to this area of research by providing empirical proof of sustainable smart PSSs with the support of AI in the form of DL-based CV models.

\subsubsection{Fundamentals of Computer Vision and Deep Learning}
CV aims to equip computers with the ability to visually perceive the world similar to humans \citep{Szeliski2010}. For decades, CV systems relied on techniques such as edge detection and filters \citep{Szeliski2010}, which are now referred to as ``traditional'' CV techniques \citep{OMahony2019}. Recently, it was shown that CV systems based on ML have the potential to produce more accurate outputs. In isolated cases, ML-based CV systems even surpassed human performance \citep{he2015delving}. 

ML is a relatively old field of research, defined in 1959 by Arthur Samuel as giving computers the ability to learn without being explicitly programmed \citep{Samuel1959}. Current CV systems are based on DL, a subfield of ML that relies on deep neural networks \citep{Janiesch2021}. 
In particular, convolutional neural networks (CNNs) are typically applied to CV tasks, as they perform well on visual data \citep{kim2017convolutional}.

In this paragraph, we briefly describe how CNNs work according to \cite{LeCun2015}. Like other types of neural networks, CNNs consist of multiple processing layers that learn to represent data with different degrees of
abstraction. However, as opposed to fully connected neural networks the same operations are applied to all inputs from the previous layer. Hence, the number of weights is
reduced considerably in comparison to fully connected networks. Consequently,
a CNN can be successfully trained with significantly less data and computing
power than a classic fully connected neural network for the same task
\citep{LeCun1995}.

\begin{figure}[htbp]
	\includegraphics[width=\linewidth]{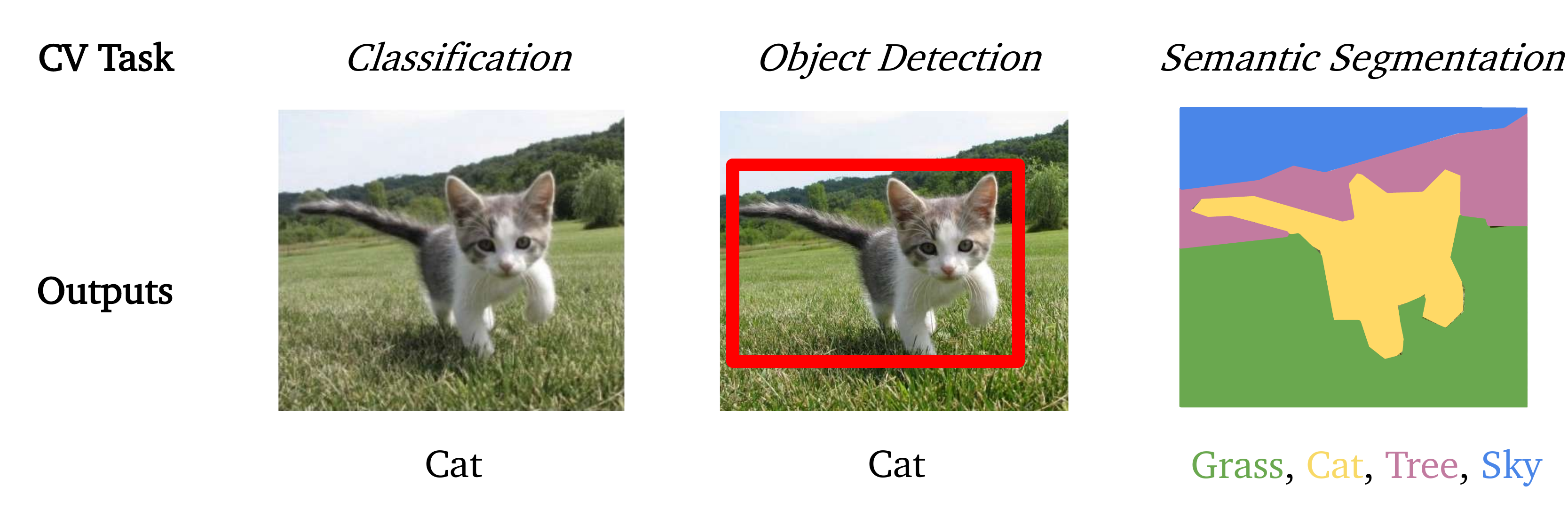}
	\caption{Overview of typical CV tasks. Our representation is based on \cite{FeiFei2017} and \cite{kossonsa}.}
	\label{CVTasks}  
\end{figure}

The following CV tasks are most typical for static images (compare \Cref{CVTasks}): \textit{ image classification}, \textit{ object detection }, and \textit{ semantic segmentation } \citep{Griebel2019a}. In image classification, the CNN output is a class label for the entire image. In object detection, a bounding box containing the object of interest is output along with the class label. The information produced by a CNN for semantic segmentation is even more fine-granular---each pixel in an input image is assigned a class label. 

\subsection{Methodology}
\begin{sidewaysfigure}
    
    \centering
    \includegraphics[width=\linewidth]{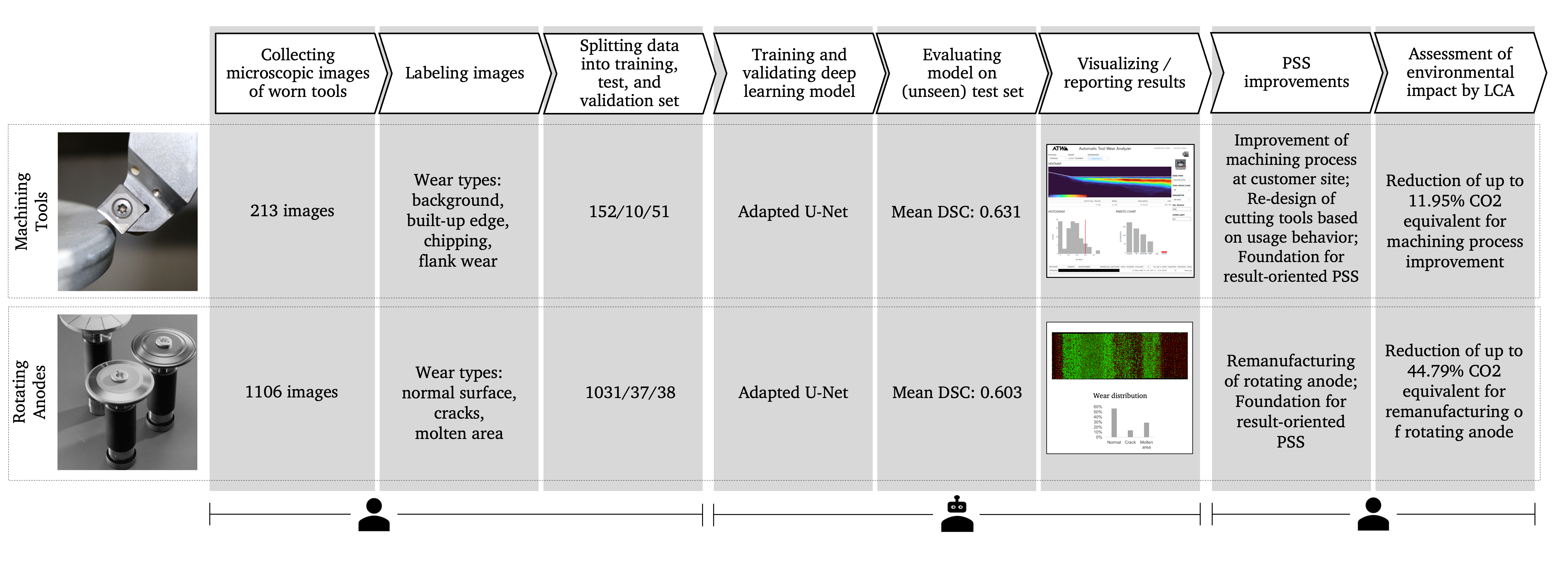}
    \caption{Overview of the methodology of this work.}
    \label{fig:Methodology}
\end{sidewaysfigure}

With the foundations of PSSs and DL-based CV models at hand, we now introduce our methodology. \Cref{fig:Methodology} presents a high-level overview of the methodology. First, we describe various steps to assess the feasibility of extracting relevant information from microscopic images using DL-based CV models. The last two steps describe the LCAs at a high level. As shown at the bottom of the figure, some steps are performed manually while others are performed by the DL-based CV models. The following subsection details the CV and LCA methodology.

\subsubsection{Computer Vision Methodology}
\label{subsubsec:ML_meth}
To allow for an accurate assessment of the wear state of products in 2D images, we train a DL-based CV model to perform semantic segmentation. Semantic segmentation provides pixel-wise information; thus, it enables a detailed assessment of the different wear types. According to the domain experts in the case companies, this level of detail is most helpful for the case studies considered. We use a CNN based on the U-Net architecture \citep{Ronneberger2015}. This architecture stems from the biomedical domain and has become a standard for different types of semantic segmentation tasks. Semantic segmentation models are typically trained in a supervised manner---the neural network learns to solve a task using the  optimal results as a (to be predicted) target during training. For semantic segmentation, the target is specified in the form of a human-created label that defines the wear class for each pixel in the corresponding image.
To train, tune, and evaluate the DL-based CV models, we split the datasets into training, validation, and test sets \citep[p. 222]{Hastie2017}. The training set is used to fit the models, the validation set is used to select a model configuration from different models with varying hyperparameters, and the test set is used to estimate the fraction of errors the DL-based CV model will commit later in a---previously unseen---real-world setting. To assess the prediction quality, we compare the human-assigned labels and predictions and compute numerical evaluation measures.
For the numerical assessment of the predictions, we rely on the pixel accuracy and mean dice coefficient (mean DSC) \citep{dice1945measures}, which is a common metric for evaluating the performance of semantic segmentation models \citep{setiawan2020image}.
It is defined as follows:
\begin{equation}
\mathit{Mean \; DSC } = \frac{2}{C} \sum_{c=1}^{C} \frac{\sum_{i=1}^{N} \hat{y}_{i,c} \; g_{i,c} }{\sum_{i=1}^{N} \hat{y}_{i,c}  + \sum_{i=1}^{N} g_{i,c}}
    \label{eq:averaged_dice}
\end{equation}
with the prediction $\hat{y}$ assigning a class label $c \in C$ to each pixel $i \in N$. $C$ represents the total number of classes and $N$ denotes the number of pixels in the input image. $g_{i,c}$ denotes the one-hot-encoded human labels used as the ground truth. We implement, train, and evaluate our model using the Python library \textit{Keras} in  version 2.1.6. \citep{chollet2015keras}.

\subsubsection{Life Cycle Assessment Methodology}
\label{sububsec:LCA_meth}
Once we are aware of the CV results, we integrate them into a user-centric artifact that supports domain experts in decision-making and facilitates the different PSSs described previously. Additionally, we discuss the CV results with domain experts from the case companies. Based on this, we examine the impact on the environmental sustainability of different PSSs together with domain experts. To assess the environmental impacts of the selected PSSs, we perform LCAs using the methodology described in the following.

LCA is an internationally standardized method used for the quantitative environmental impact assessment of products, processes, services, and systems throughout their life cycles \citep{finkbeiner2006new}. LCA can be particularly useful for comparing alternative strategies and understanding the trade-offs between the benefits and impacts of different systems, which can help in making informed decisions \citep{niero2014comparative}. LCA can be deployed as a quantitative decision-support tool in sustainable design engineering or green manufacturing \citep{saidani2021comparative}. 

According to ISO standards 14040 \citep{ISO14040} and 14044 \citep{ISO14044}, an LCA comprises four major steps:
\begin{enumerate}
    \item Goal and scope definition: The goal phase defines the overall objectives of the study. The scoping phase sets the boundaries of the system studied, sources of data, and functional unit to which the results refer.
    \item Life cycle inventory (LCI): A detailed compilation of all the inputs (e.g., material and energy) and outputs (e.g., pollutants) at each stage of the life cycle is performed.
    \item Life cycle impact assessment (LCIA): It aims to quantify the relative importance of all environmental burdens obtained in the LCI by analyzing their influence on the selected environmental impact categories.
    \item Interpretation of results: The outcomes of the LCI and LCIA stages are interpreted to find hotspots and compare alternative scenarios.
\end{enumerate}
\newpage
A key aspect to consider in the goal and scope definition is the functional unit (FU). It provides a reference to which the inputs and outputs of the LCA can be related \citep{cooper2003specifying}. According to ISO 14044  \citep{ISO14044}, the FU should be clearly defined and measurable. This enables a scientifically sound (i.e., consistent and unbiased) comparison between different product systems and scenarios. \citet{JointResearchCenter2010} recommends including the following aspects in the definition of the FU: (i) verb (functional analysis); (ii) what (form of the output); (iii) how much? (magnitude), how well? (performance), and how long? (duration).

For the LCIA, the OpenLCA software (2021 version 1.10.3), developed by GreenDelta \citep{ciroth2014openlca}, was used to model the PSSs and conduct comparative LCAs. Within OpenLCA, the Ecoinvent database (2021 version 3.7) \citep{wernet2016ecoinvent} and the ReCiPe 2016 Midpoint (H) method were used to perform the environmental evaluation. Ecoinvent is one of the most comprehensive and acknowledged databases providing the necessary data for impact calculations, including region-specific production and manufacturing data for numerous commodities across multiple industries \citep{frischknecht2005ecoinvent}. ReCiPe is a scientifically sound and acknowledged impact calculation method that provides characterization factors and normalization methods for calculating the impact \citep{huijbregts2017recipe2016}.

Note that in the present case, the application of LCA appears at the end of our methodology (see \Cref{fig:Methodology}) as LCA is used, here, to quantify the difference of environmental impacts between the baseline scenarios (without the DL-based CV model) and the smart ones (PSSs augmented with DL-based CV model). In practice, in the present case (i.e., through the two application cases), the improvement scenarios are driven by the new capabilities enabled by the DL-based CV models. On this basis, comparative LCAs are performed to validate the environmental benefits of adding DL-based CV models to the systems, factoring in the energy required to train the DL-based CV models (as detailed in \Cref{subsec:results_LCA}). In fact, while PSSs have the potential to improve environmental sustainability and enhance customer satisfaction \citep{fargnoli2018product}, they are not more sustainable by default. Consequently, environmental assessment tools, such as LCA, are essential to validate the environmental performance of PSSs \citep{kjaer2016challenges}.
\newpage
\subsection{Case Studies}
In this subsection, we introduce our case studies: the products, necessary domain knowledge, and respective industries. In each case study, we introduce DL-based CV models as a tool to better assess wear and, subsequently, demonstrate how the results impact environmental sustainability.
\subsubsection{Machining Tools}
\label{subsubsec_case_machining_tools}
\begin{figure}
    \centering
    \includegraphics[width = 0.5\linewidth]{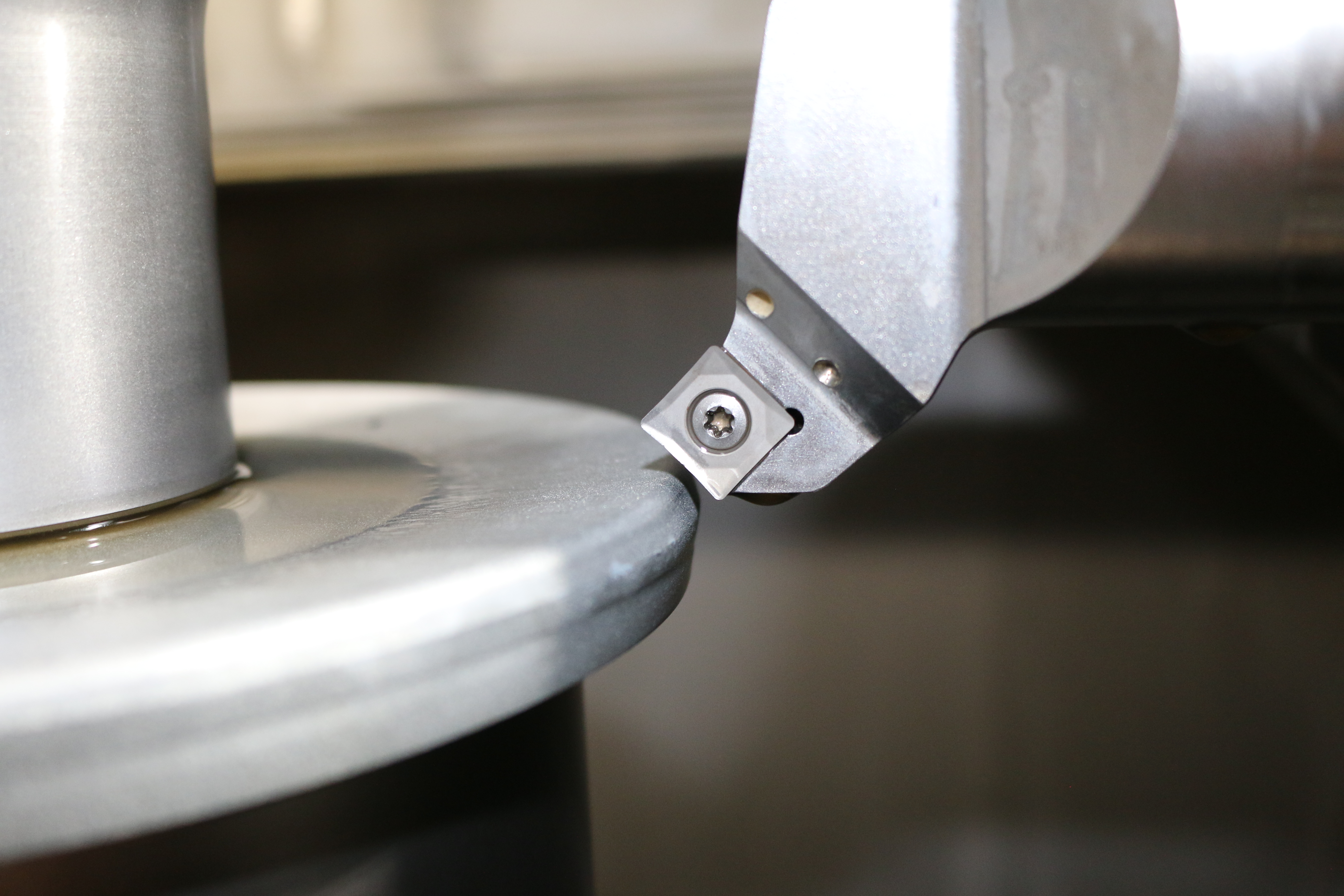}
    \caption{Exemplary turning process.}
    \label{fig:TurningProcess}
\end{figure}

Machining is an important manufacturing process \citep[p. VI]{Black1995}. It is utilized in numerous industries, such as healthcare \citep{churi2009rotary}, aerospace \citep{ezugwu2005key}, and automotive \citep{tai2014minimum}. \Cref{fig:TurningProcess} shows an exemplary turning process, a specific type of machining process. During the turning process, the workpiece rotates at a high speed; consequently, there is a relative motion between the workpiece (left) and the cutting tool (right). This results in the removal of unwanted material from the workpiece \citep[p. VI]{Black1995}.
Owing to the powerful forces and elevated temperature, the cutting tool is subject to wear and must be changed regularly \citep{bergs2020digital}. Analyzing the wear of the cutting tool is essential for understanding the improvement potential of the machining process. Because the cutting tools are small and frequently changed, it is economically unviable to equip the tools themselves with sensors for connectivity, as is the case with several other small tools with low unit prices \citep{martin2019holistic}. Consequently, the visual inspection of worn cutting tools is an important building block for understanding the wear and, therefore, the improvement potential. 

Machining processes are influenced by several interdependent components such as cutting tools, tool holders, workpieces, workpiece holders, engines, and cutting fluids. The interplay of these components leads to an inherent variance in the machining processes. Particularly, the wear on two tools used in an identical process can vary significantly. Analyzing a single tool only provides a snapshot of the entire process. By contrast, analyzing several tools provides a more holistic overview of a given machining process. However, manual visual inspection of cutting tools requires considerable effort and, therefore, is currently not performed on a large scale. Visual inspection by CV facilitates efficient analysis of a large number of cutting tools and consequently allows for more reliable conclusions regarding the machining process. As described in \Cref{subsubsec:ML_meth}, we utilize a DL-based CV model to detect the wear on worn cutting tools. 

The possible sustainable smart PSSs resulting from this can be grouped into \textit{improvement of machining processes at customer sites}, \textit{re-design of cutting tools based on behavior in use}, and \textit{establishing the foundation for result-oriented PSSs}. These three approaches will be discussed in greater detail in the following section in terms of the improvement of environmental sustainability.

Tool producers typically have teams of specialized application engineers who support customers in improving their machining processes. The visual inspection of worn cutting tools is an important part of their job. Domain experts describe the current process for machining process improvements as follows: Typically, an application engineer visits the production site of a customer and inspects a small number of worn cutting tools (e.g., three) with a magnifier. Based on this, as well as additional information, such as machining parameters and domain expertise, 
a recommendation for process improvement is provided and implemented. Our approach enables the inspection of many worn cutting tools as a decision basis. Owing to its efficiency, it is possible to assess the wear state of a large number of worn cutting tools, for example, 200. This provides more reliable insights into the process improvement potential. Consequently, better results are expected from the process improvements. Additionally, the efficiency of the DL-based CV model enables more process improvements. In \Cref{subsubsec:LCA_machining}, we present the results of an LCA for machining process improvements based on the proposed DL-based CV model for wear assessment.

These insights can also be utilized in a more strategic way---understanding the usage behavior of machining tools can aid the design of the next generation of machining tools. Currently, the development of new generations of machining tools relies mainly on internal tests in controlled environments, which do not necessarily reflect real usage and the possible variances therein.

\begin{figure*}[ht] 
    \centering
  \begin{subfigure}[b]{0.28\linewidth}
    \centering
    \includegraphics[width=1\linewidth]{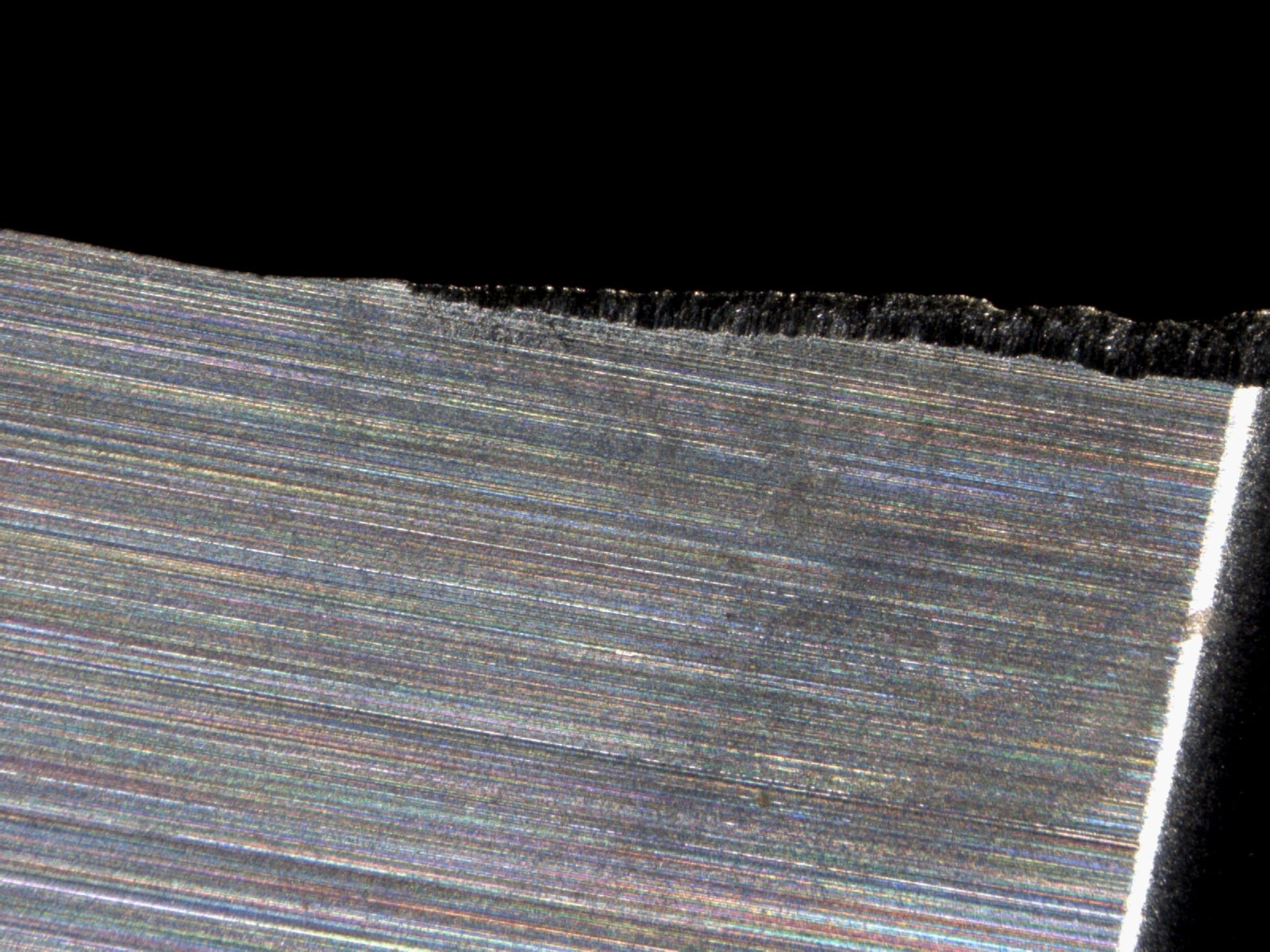} 
    \caption{Flank wear} 
    \label{flank_wear} 
  \end{subfigure}
  \hfill
  \begin{subfigure}[b]{0.28\linewidth}
    \centering
    \includegraphics[width=1\linewidth]{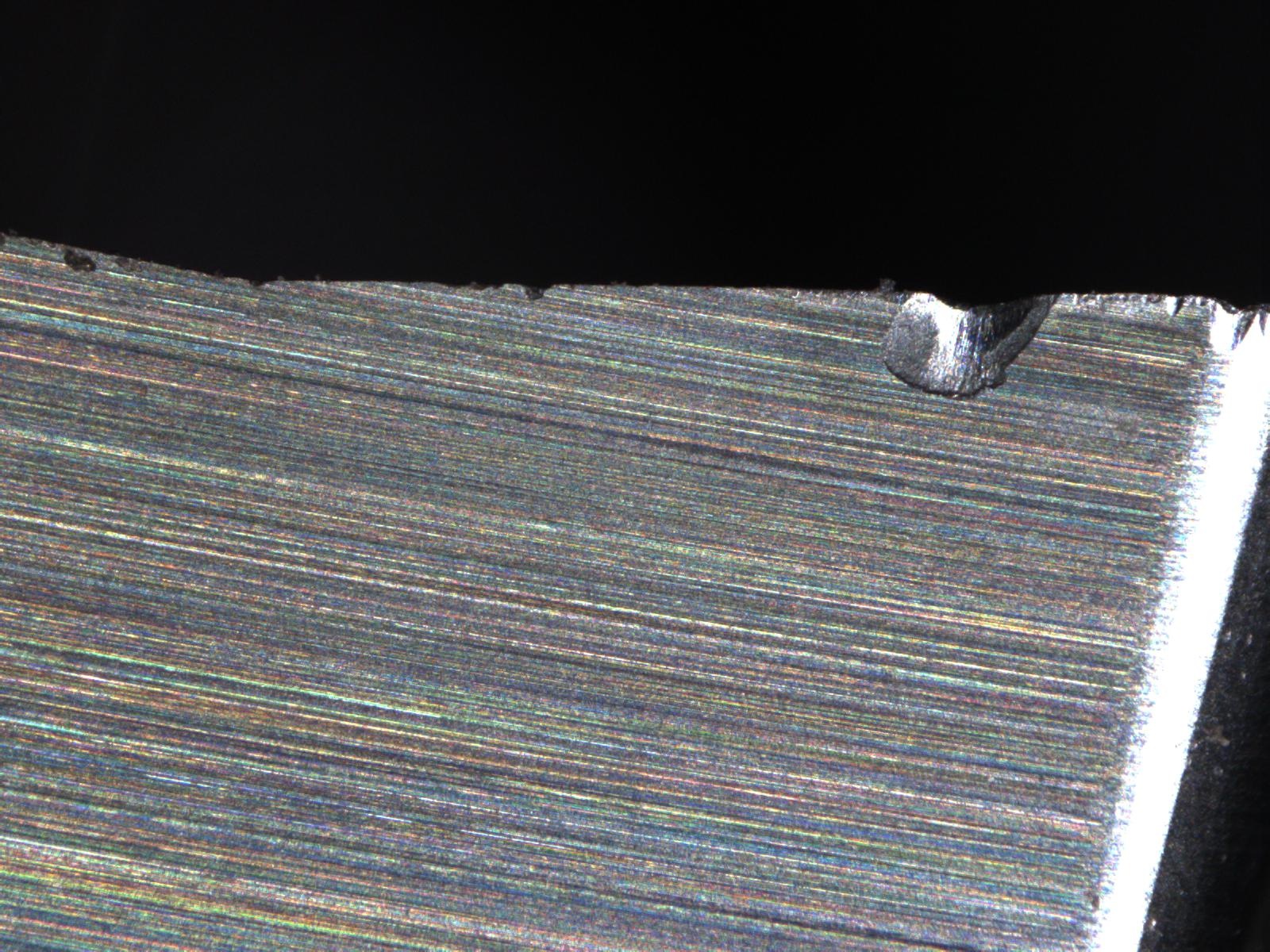} 
    \caption{Chipping} 
    \label{chipping} 
  \end{subfigure} 
  \hfill
  \begin{subfigure}[b]{0.28\linewidth}
    \centering
    \includegraphics[width=1\linewidth]{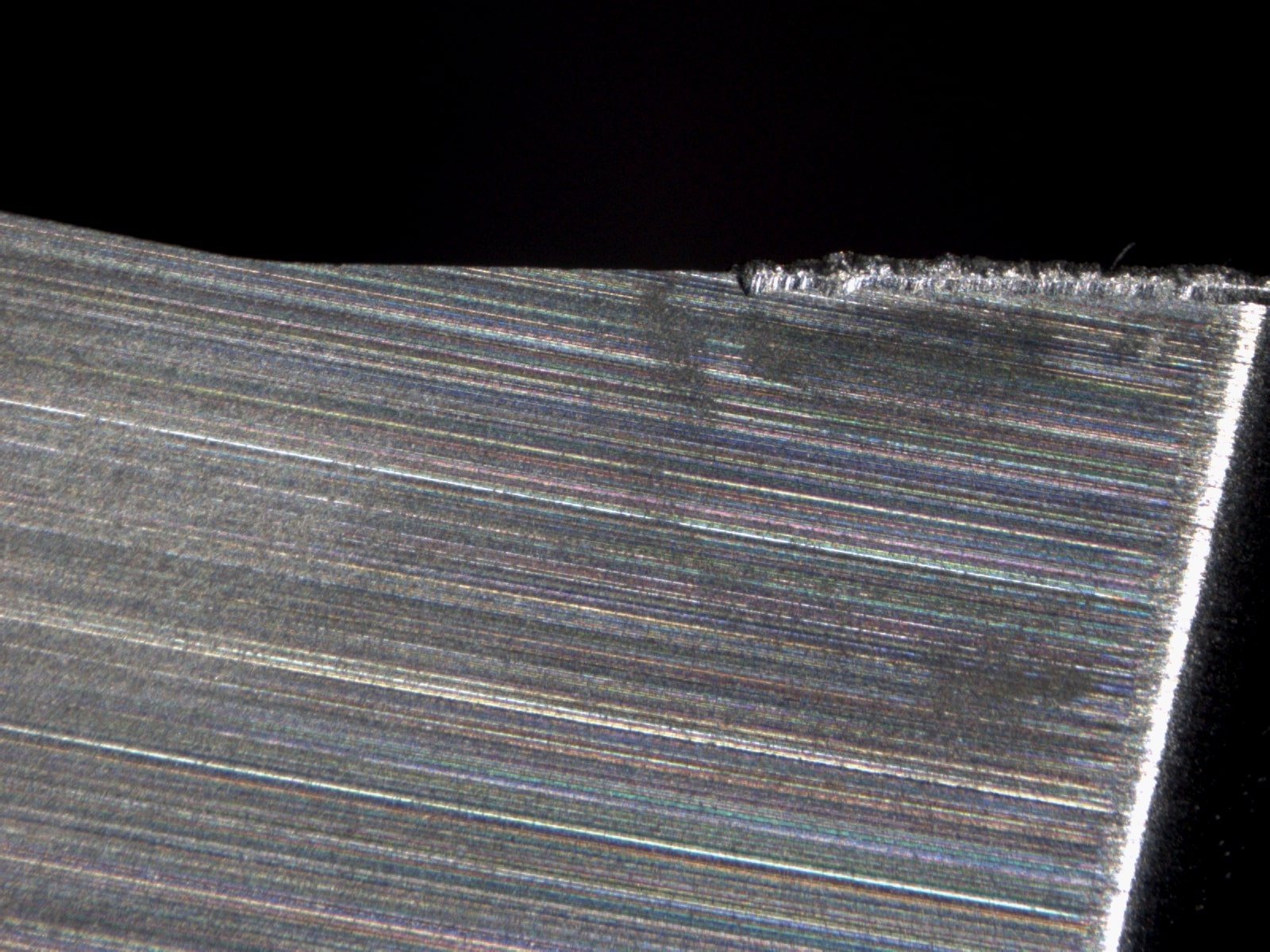} 
    \caption{Built-up edge} 
    \label{built-up-edge}
  \end{subfigure}
    \hfill
  \caption{Common wear mechanisms in machining processes.}
  \label{fig:examined_wear} 
\end{figure*}

Ultimately, understanding the usage behavior of machining tools can support a change in the business model. Tool manufacturers typically sell their products to companies that use them in their production. Research suggests that PSSs, particularly result-oriented ones, can yield both environmental sustainability \citep{Tukker2015} and economic benefits \citep{Annarelli2016, yang2019product}. A major hurdle for the adoption of PSSs is the difficulty of the offering party in estimating expected costs and risks \citep{Erkoyuncu2011}. Visual inspection of worn cutting tools using CV can facilitate a quick and accurate estimation of the risks and costs for a given machining process.
\newpage
For this case study, we collected 213 worn cutting inserts from a real production process and captured microscopic images of the cutting edges. The wear mechanisms relevant to this case study are shown in \Cref{flank_wear} -- \Cref{built-up-edge}. \textit{Flank wear} results from friction between the cutting tool and workpiece \citep{Altintas2012}. This is the preferable wear mechanism because it occurs continuously. Also, it is the most frequent wear mechanism \citep{siddhpura2013review}. In contrast, \textit{chipping} and \textit{built-up edge} are less desirable because they occur suddenly, leading to a significant deformation of the cutting edge. The cutting edge deformation can lead to the workpiece being considered scrap owing to its poor surface quality. \textit{Chipping} describes the phenomenon of the cutting edge particles breaking off. A \textit{built-up edge} occurs because of stress in the form of heat and pressure, resulting in the deposition of workpiece material on the cutting edge. In our case study, we trained a DL-based CV model to detect these wear mechanisms using microscopic images of worn cutting edges.
A detailed description of the ML-related technical details of this case study can be found in \cite{treiss2020uncertainty}.

We closely cooperated with the domain experts of the case company to validate the results of the CV model and LCA. Additionally, we were in frequent contact with them to obtain real data and assumptions as inputs for LCAs. \Cref{ExpertsMachiningCase} describes the domain experts as well as the IDs used for the remainder of this work.

\begin{table}[]
\caption{Domain experts for the machining tools case.}
\label{ExpertsMachiningCase}
\centering
        \begin{tabular}{lcc}
        \toprule
            ID & Role & \makecell{ experience \\in years} \\
            \midrule
            Alpha 	&
            development engineer &
            5-10\\
            Beta 		  &   \makecell{ development engineer} &
            5-10\\
            Gamma		   &   application engineer &
            \textless5\\
            Delta &  
            application engineer
            &
            \textgreater10\\
            \bottomrule
        \end{tabular}
\end{table}

\subsubsection{Rotating Anodes}
\label{subsubsec_case_XRT}
\begin{figure}
    \centering
    \includegraphics[width = 0.45\linewidth]{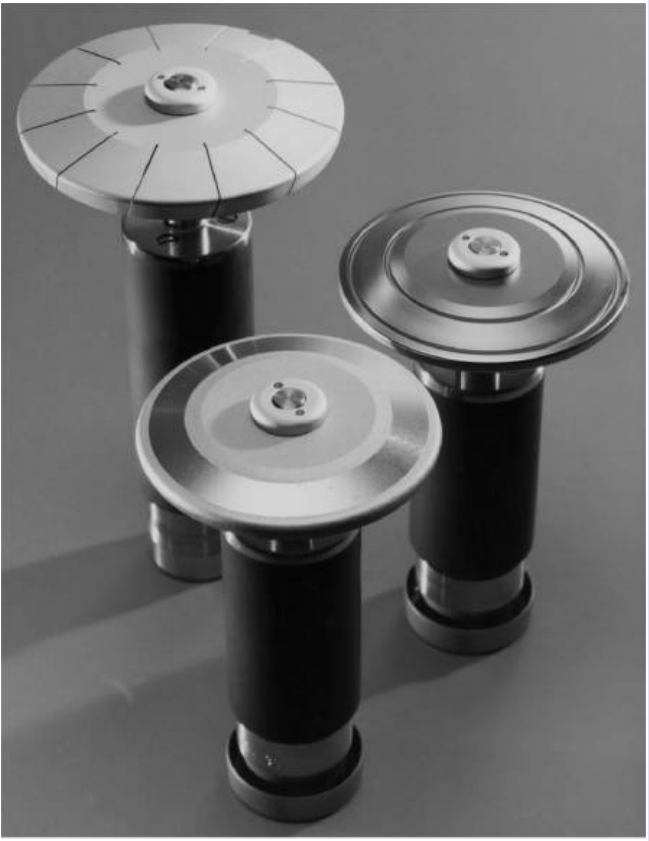}
    \caption{Rotating anodes based on \cite{behling2015modern}.}
    \label{fig:rotating_anodes}
\end{figure}

\begin{figure}
    \centering
    \includegraphics[width = 0.45\linewidth]{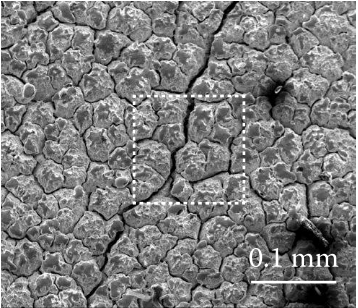}
    \caption{Microscopic image of eroded rotating anode focal track from \cite{behling2015modern}.}
    \label{fig:focal_track}
\end{figure}

Modern medicine relies on X-rays for the diagnosis of injuries and illnesses, for example, broken bones \citep[p. 110]{behling2015modern}\footnote{Please note that rotating anodes are a niche and specialized topic. The reference books (and largely the only ones) for this topic are \cite{behling2015modern} and \cite{oppelt2005imaging}, which is why they are frequently used in this section. The rights for the reuse of images (\Cref{fig:rotating_anodes} and \Cref{fig:focal_track}) are granted.}, breast cancer \citep[p. 139]{behling2015modern}, and cardiac and vascular diseases \citep[p. 479]{oppelt2005imaging}. To produce X-rays, a cathode emits electrons via thermal emission. These electrons are accelerated towards an anode, where they are decelerated, and X-rays emerge as a result \citep[p. 180]{behling2015modern}. Rotating anodes are typically used for high-intensity X-rays \citep[p. 18]{behling2015modern}. The rotation counteracts overheating because the electron beam from the cathode hits different spots along the circumference of the anode \citep[p. 283]{oppelt2005imaging}. The rotating anode can be considered the most important part of an X-ray tube, and is usually one of the most expensive parts. It determines the performance of the overall system (\cite[p. 233]{behling2015modern}, \cite[p. 280]{oppelt2005imaging}). During operation, temperatures of rotating anodes reach up to 1,500 °C (2,732 °F) \citep[p. 240]{behling2015modern} with microsecond-long pulses of up to 2,500 °C (4,532 °F) \citep{Mehranian2010}. Consequently, in the area where the electron beam hits the rotating anode, the focal track erodes due to extreme thermal cycling \citep[p. 235, 247]{behling2015modern}. Focal tracks show two types of wear \citep[p. 248]{behling2015modern}:  \textit{cracks} occur because of repeated heating and cooling, and they provide stress relief. \textit{Molten areas} appear because the grains of the focal track are isolated due to cracks, and, consequently, heat conduction to the surrounding material is limited. The focal track wear often limits the lifespan of a rotating anode \citep{Erdelyi2009}.  Examples of rotating anodes are shown in \Cref{fig:rotating_anodes}. \Cref{fig:focal_track} shows a microscopic image of an eroded focal track. 

The state of the focal track often limits the lifespan of the rotating anode.
We utilize a DL-based CV model to detect and quantify the wear state of the focal tracks of the worn rotating anodes. In this case, it is not only more scalable and reproducible to characterize the wear state of the product. Owing to the large size of the rotating anodes and focal tracks and the small size of the wear mechanisms, it is almost impossible for humans to assess the wear state in detail. During our project, we captured microscopic images of the focal tracks, which enable an assessment of the wear state. A single microscopic image is more than 19,000 pixels high and 5,000 pixels wide. This microscopic image shows less than 0.5\% of the entire focal track.\enlargethispage*{0.5cm} Hence, a detailed manual assessment of the wear state of an entire focal track is tedious. 

The possible sustainable smart PSSs resulting from this detailed assessment of the wear state by CV can be grouped into \textit{remanufacturing and recycling decisions, re-design of rotating anodes based on actual wear}, and \textit{establishing the foundation for result-oriented PSSs}.
In this section, we further describe these three approaches for improving environmental sustainability.

An assessment of the wear state of the focal track of a rotating anode using CV enables making an efficient and reproducible decision regarding the options remanufacturing and recycling. There are different options for remanufacturing depending on the type and severity of the wear. If none of the remanufacturing options is applicable, the rotating anode must be recycled. As described by \citet{fang2015use}, uncertainty regarding the state of returned products is a major hurdle in remanufacturing endeavors, which is addressed by our approach. In \Cref{subsubsec:LCA_XRT}, we present the results of an LCA that compares a baseline scenario with a remanufacturing scenario.

Similar to machining tools, the development of new generations of rotating anodes can be aided by actual usage behavior and wear instead of internal laboratory settings.

The rationale behind wear information serving as the foundation for establishing result-oriented PSSs is also in accordance with that regarding the machining tools described in \Cref{subsubsec_case_machining_tools}.

For this case study, we collected images of focal tracks of several worn rotating anodes that were sampled for a realistic wear distribution. As described before, the images captured by the microscope are large; therefore, labeling an entire image manually would be extremely tedious. Instead, we choose 1,106 representative small patches of microscopic images together with domain experts, labeled them, and used them for the training and evaluation of our DL-based CV model. 

As stated previously, we cooperated closely with the domain experts of the respective case company. In addition to selecting representative image patches, they provided input data for the LCA and validated the results of the CV model and LCA. \Cref{ExpertsXRTCase} describes the domain experts as well as the IDs we use for the remainder of this work.
\begin{table}[]
\caption{Domain experts for the rotating anode case.}
\label{ExpertsXRTCase}
\centering
        \begin{tabular}{lcc}
        \toprule
            ID & Role & \makecell{ experience \\in years} \\
            \midrule
            Epsilon 	&
            \makecell{development engineer} &
            5-10\\
            Zeta 		  &   \makecell{ development engineer} &
            \textgreater10\\
            Eta 		  &   \makecell{sales manager} &
            \textgreater10\\
            Theta 		  &   \makecell{sales manager} &
            \textless5\\
            \bottomrule
        \end{tabular}
\end{table}

\section{Results}
\label{sec:Results}
In this section, we report the application results of our previously described methodology. More precisely, we first describe the CV results in \Cref{subsec:CV_results} and subsequently in \Cref{subsec:results_LCA}, the results of the LCAs.

\subsection{Computer Vision Results}
\label{subsec:CV_results}
In the following, we describe the results of the DL-based CV models for the machining tools and rotating anode case study.
\subsubsection{Machining Tools}
\begin{figure}[ht] 
      \label{performance_results_tool_wear}
      \centering
  \begin{subfigure}[b]{0.75\linewidth}
    \centering
    \includegraphics[width=0.95\linewidth]{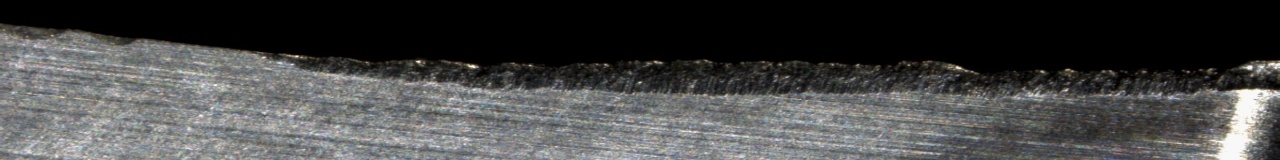} 
    \caption{Preprocessed input image.} 
    \label{fig_seg_map_a} 
  \end{subfigure}
  \vspace{\floatsep}
  \begin{subfigure}[b]{0.75\linewidth}
    \centering
    \includegraphics[width=0.95\linewidth]{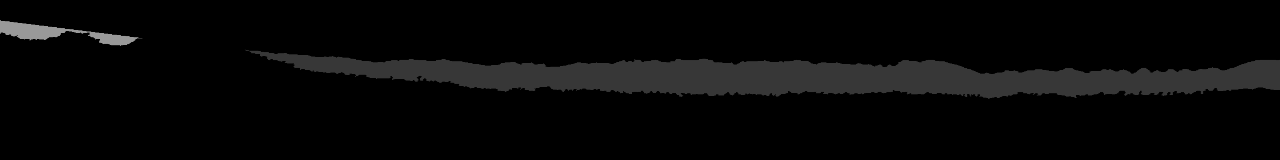} 
    \caption{Human label.} 
    \label{fig_seg_map_c}
  \end{subfigure}
  \vspace{\floatsep}
  \begin{subfigure}[b]{0.75\linewidth}
    \centering
    \includegraphics[width=0.95\linewidth]{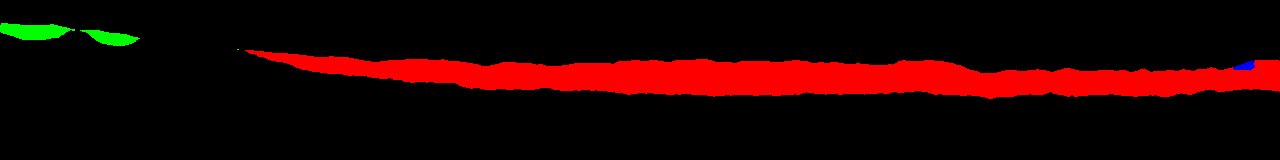} 
    \caption{Prediction.} 
    \label{fig_seg_map_e} 
  \end{subfigure}
  \caption{ Image from test set with the corresponding human label and prediction of the neural network (best viewed in color).\\ Color coding: flank wear = dark gray/red, chipping = light gray/green, and built-up edge = white/blue.}

  \label{fig_seg_map} 
\end{figure}

We split our dataset containing 213 microscopic images of worn cutting tools into 152 training images, 10 validation images, and 51 test images. \Cref{fig_seg_map} shows an exemplary original image, human label, and prediction from the test set. 

\Cref{ToolWear_performance_results} lists the performance results of our trained U-Net on the unseen test set. Theoretically, the DSC can reach values between zero (no overlap between prediction and ground truth) and one (perfect overlap). In our case, depending on the wear type, we obtained results between 0.244 (chipping) and 0.991 (background). We attribute the relatively poor prediction performance for chipping to two phenomena: first, chipping seldom occurs in our dataset. Consequently, there is relatively little data from which the model can learn. Second, the DSC drops to zero in the case of a false positive, that is, the respective wear class is predicted but not present in the ground truth, which occurs several times for chipping in our dataset. We obtained a mean DSC value of 0.631. Although there is no unified scale to judge the results, they are in accordance with related work in the medical domain \citep{zou2004statistical}. \cite[p. 51]{guindon2017application} state that a DSC of 0.7 is ``deemed to be indicative of an excellent match between the segmentation result and human expert delineation''. However, to further validate the results, we engaged in a dialogue with domain experts from our case company. During the discussions, they confirmed the feasibility of our approach (Alpha and Beta).

\begin{table}
\caption{Performance results of the machining tools dataset.}
\label{ToolWear_performance_results}
\centering
        \begin{tabular}{lc}
        \toprule
            Class and {Dice coefficients } \\
            \midrule
            Background  	& 	0.991\\
            Flank wear 		  &   0.695	\\
            Chipping 		   &   0.244	\\
            Built-up edge   &    0.596 \\
            &  \\
            \midrule
            \midrule
            Mean DSC &  0.631 \\
            Pixel accuracy 		  & 0.977 \\ 
            \bottomrule
        \end{tabular}
\end{table}

\begin{figure}
    \centering
    \includegraphics[width = 1\linewidth]{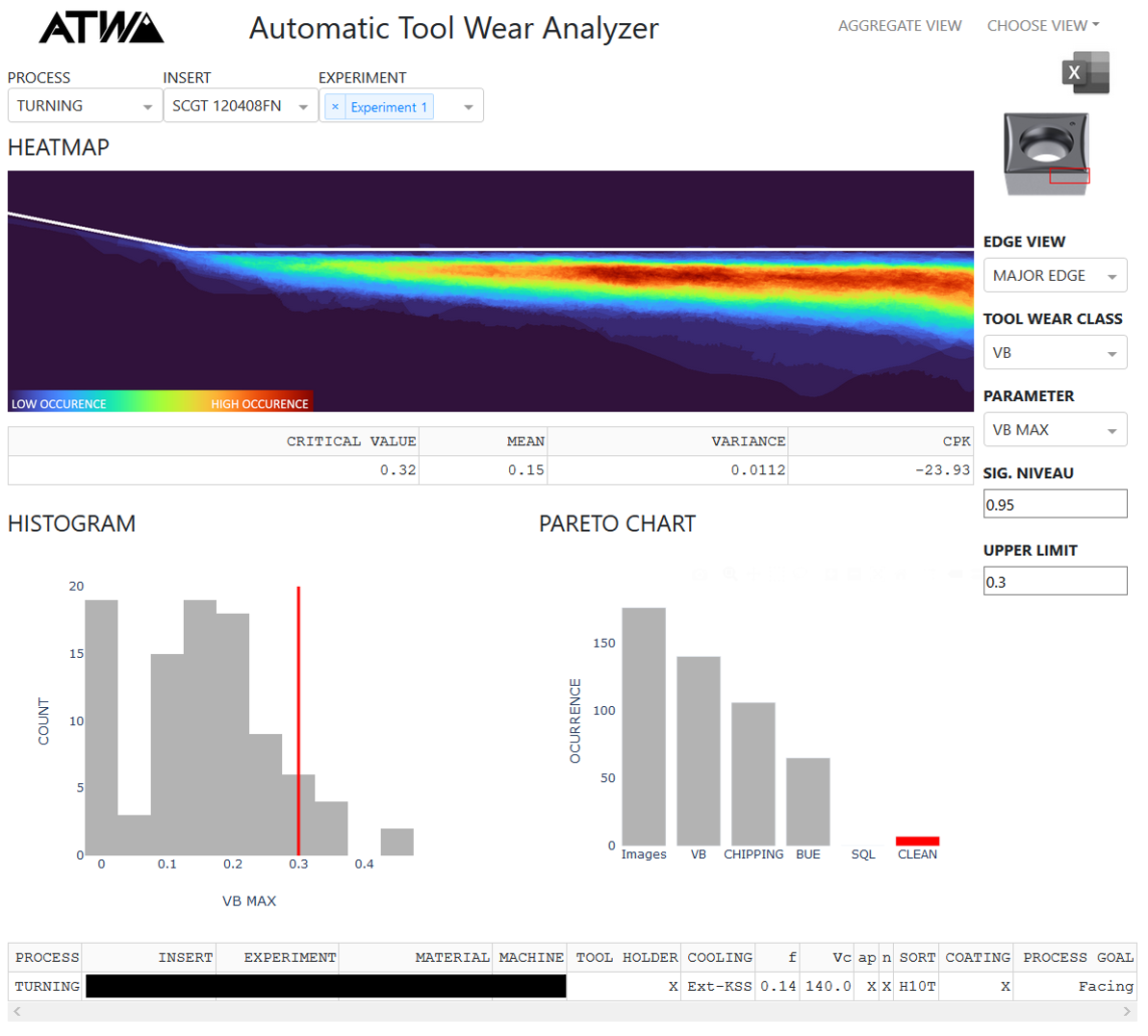}
    \caption{Exemplary screenshot of user-centric artifact for machining tool wear interpretation.}
    \label{fig:ATWA_Overview}
\end{figure}

Although an efficient assessment of the wear state of machining tools is already helpful for domain experts, the utility is further increased by the integration of the outputs in a user-centric artifact. A screenshot showing an exemplary view of the artifact is shown in \Cref{fig:ATWA_Overview}. 
This user-centric artifact was developed iteratively with seven domain experts as end users. It visually and statistically aggregates a dataset consisting of images of worn cutting tools from the same process assessed using CV. Overall, it enables domain experts to interactively explore the dataset. Consequently, they can combine the information extracted through CV with their domain expertise to make data-driven decisions. For example, an application engineer can use the user-centric artifact based on the results of the DL-based CV model for wear detection to optimize the machining processes of customers.

\subsubsection{Rotating Anodes}
We split our dataset of 1,106 microscopic image patches into 1,031 training images, 37 validation images, and 38 test images.
The results reach performance levels comparable to those of the machining tools case study. These results are listed in \Cref{XRT_performance_results}. They demonstrate the feasibility of extracting relevant information from images using CV according to domain experts (Epsilon and Zeta). In direct comparison with the machining tools, the somewhat lower values for the evaluation metrics can be explained by the fact that the images of focal tracks of the rotating anodes do not contain any black background, which is particularly easy to detect.

\begin{table}[]
\caption{Performance results of the rotating anodes dataset.}
\label{XRT_performance_results}
\centering
        \begin{tabular}{lc}
        \toprule
            Class and {Dice coefficients } \\
            \midrule
            Normal surface 	& 	0.485\\
            Cracks 		  &   0.634	\\
            Molten area		   &   0.690	\\
            \midrule
            \midrule
            Mean DSC &  0.603 \\
            Pixel accuracy 		  & 0.737 \\ 
            \bottomrule
        \end{tabular}
\end{table}
A similar user-centric artifact is conceptualized for rotating anodes, which allows domain experts to explore the wear state as detected by CV. Domain experts can use it to better understand the wear of rotating anodes---a prerequisite for designing new generations of rotating anodes based on the usage behavior and for offering result-oriented PSSs.

 \subsection{Life Cycle Assessments}
\label{subsec:results_LCA}
\enlargethispage*{1cm}
Using the performance metrics of the CV models, the potential impact savings and environmental effects owing to their usage within the processes at our case companies can be calculated. We followed the LCA terminology depicted in \Cref{sububsec:LCA_meth}. 

\textbf{Goal and scope definition:} The main objective of the current LCAs is to assess the potential impact savings facilitated by the DL-based CV models to enhance the usage and lifespan of manufactured products through PSSs. For the two case studies, the status-quo situation is compared with improvement scenarios that are supported by the DL-based CV models:

\begin{itemize}
    \item For the machining tools case, the FU is defined as follows:  Manufacture 100 unit shafts (42CrMo4, 800 grams) with tungsten carbide cobalt cutting tools (WC-Co, 9.06 grams) according to predefined specifications.
    \item For the rotating anode case, the FU is defined as follows: Provide two rotating X-ray anodes according to predefined specifications for assumed usage of five years.
\end{itemize}
The system boundaries for the two case studies are as follows:
\begin{itemize}
    \item For the machining tools case, the cutting tool, electricity for the machining center, cutting fluid, and CV model for wear assessment are considered within the scope of the LCA.
    \item For the rotating anode case, two rotating anodes are within the scope of the LCA. Depending on the scenario, either both are produced from scratch, or the second is remanufactured. Additionally, transportation from the production site to customers (and back if necessary) is considered. Additionally, the CV model for wear assessment is accounted for.
\end{itemize}

\textbf{Life cycle inventory (LCI):} Product factsheets and real measurements provided by our case companies have been used for the LCI whenever possible, e.g., for the exact material composition of the cutting tool, and the energy and resource consumption of several production steps of rotating anodes. If this was not possible, assumptions made by the domain experts described in \Cref{ExpertsMachiningCase} and \Cref{ExpertsXRTCase} were used in combination with the scientific literature on the respective topics. 

For the machining tools case, the expected lifespan under the baseline scenario of the cutting tool is estimated to be 30 minutes of operation (Gamma and Delta). The manufacturing of one unit shaft is estimated to take 30 seconds. Considering breaks for the employee and setup times, we assume that 100 unit shafts can be produced in one hour under the baseline scenario (Gamma and Delta). The energy consumption of the machining center is estimated to be 12.5\,kWh (Gamma and Delta) on average, using the German electricity mix, that is, where the pieces are manufactured in the present case. Regarding the lubrication system, the consumption of cutting fluid is 0.0155 liters per hour, based on real data from a customer of our case company. Finally, the actual surplus energy consumption due to the training of the DL-based CV model for wear assessment is considered and detailed in the improvement scenarios to account for potential impact transfers \citep{bonvoisin2014integrated}.

The considered improvement scenarios are improvements in machining processes at customer sites by application engineers using the DL-based CV model for wear assessment. We make the (rather conservative) assumption that one trained model can be used to manufacture 1000 unit shafts. In theory, the CV model can be used an infinite number of times for the same type of machining process. The electricity usage to train and run the CV model and run the user-centric artifact is estimated to be 2.4\,kWh, as detailed in the Appendix. 

The improvement of machining processes usually aims at increasing both the lifespan of the cutting tool and the process speed, while still meeting the predefined specifications. In the following, typical machining process improvement scenarios for our FU are described based on the experience of our case company's application engineers. According to the domain experts (Alpha, Beta, Gamma, and Delta), an efficient assessment of the wear state can support these process improvements by providing accurate and fine granular information, and can hence lead to better outcomes. Additionally, the domain experts (Alpha, Beta, Gamma, and Delta) confirmed that the wear assessment by CV is highly efficient because little manual effort is required. Consequently, it is possible to provide additional process improvements to customers.
A typical machining process improvement for our FU enables an enhanced lifespan of 20\% on average for the present cutting tool, and increases the speed by 20\% on average with a maximum increase of up to 50\%. Note that there exists a clear trade-off between the cutting speed and lifespan. Higher cutting speeds result in shorter tool lifespans. For this process, it is estimated that increasing the cutting speed by 20\% decreases the tool lifespan to 70\% of its original lifespan, and increasing the cutting speed by 50\% decreases the tool lifespan to 30\%, based on experience from internal tests (Beta, Gamma, and Delta) and existing literature (e.g., \cite{konig2008fertigungsverfahren}). Despite this trade-off relationship, domain experts (Gamma and Delta) confirm that it is often possible to achieve a longer lifespan and higher cutting speed. Based on this, the improvement scenarios are computed as follows: 

\begin{itemize}
    \item Lifespan increased by 20\%
    \item Speed increased by 20\% (implies more wear and tear on the tool as aforementioned, i.e., the tool needs to be replaced more often but less electricity and cutting fluid consumption for the same FU)
    \item Speed increased by 50\%
    \item Lifespan increased by 20\% and speed increased by 20\%
    \item Lifespan increased by 20\% and speed increased by 50\%
\end{itemize}

For the rotating anode case, we work with a typical rotating anode that weighs 1.9 kilograms and comprises 12.5\% tungsten-rhenium alloy (a typical 95\% tungsten and 5\% rhenium mix \citep[p. 284]{oppelt2005imaging}) for the focal track, 12.5\% graphite for the metallic disc, and 75\% molybdenum for the cup. For this LCA, we explicitly consider the energy and resource consumption of the production steps because they have a high impact. 

The lifespan of the rotating anode is estimated by experts (Eta and Theta) to be on average 2.5 years. Because the FU refers to five years of operation, two new rotating anodes are required in the baseline case, which corresponds to the current predominant usage in this industry. As described in \Cref{subsubsec_case_XRT}, the state of the focal track often limits the lifespan of the rotating anode \citep{Erdelyi2009}. Different remanufacturing strategies can be applied to restore the focal track depending on its wear state. Consequently, the rotating anode has another 2.5 years of estimated lifespan (Eta and Theta). The CV model for wear assessment is crucial for determining whether remanufacturing is possible and the suitable strategy.

For the LCAs of the different scenarios, we almost always consider real measurements of, for example, the energy consumption of different production steps. If such values are not available, we rely on the assumptions made by the domain experts of our case company. Additionally, for this LCA, we considered the electricity consumption of the DL-based CV model, which is estimated to be 2.875\,kWh (description of this can be found in the Appendix).

To factor in transportation impacts, we consider distances from our case company in Austria to representative customers in Europe (874 km on trucks), and Asia and the United States (124 km on trucks and on average 8930.5 km by plane). Rotating anodes are typically recycled close to their last usage sites. Consequently, in the baseline case, we assume only a one-way trip from the production site to the respective customers. In the remanufacturing scenario, a round trip from the customer to the production site and back is considered. The transportation phase is represented by ton-kilometers (tkm), which is defined as the transport of 1 ton of material over a distance of 1 km \citep{Goedkoop2008}. 

Note that the impacts from the infrastructure needed to support the manufacturing facilities are beyond the scope of this study and therefore not included in the LCAs of both case studies. 

\subsubsection{Life Cycle Assessment: Machining Tools}
\label{subsubsec:LCA_machining}
For the machining tools case study, we first compare the carbon footprint of the baseline with the DL-based CV-supported improvement scenarios. We then expand our analysis to the 18 ReCiPe midpoint indicators to consolidate our interpretation and/or fine-tune our recommendations, for example, in the case of impact transfers. 

\begin{figure}[htbp]
	\includegraphics[width=\linewidth]{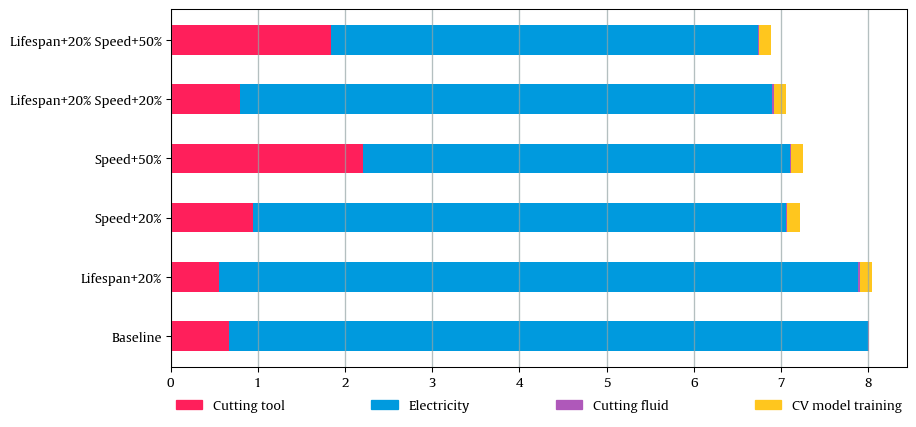}
	\caption{Carbon footprint for multiple scenarios in the machining case.}
	\label{MachinigGlobalWarming}
\end{figure}

\begin{sidewaysfigure}[htbp]
	\includegraphics[width=\linewidth]{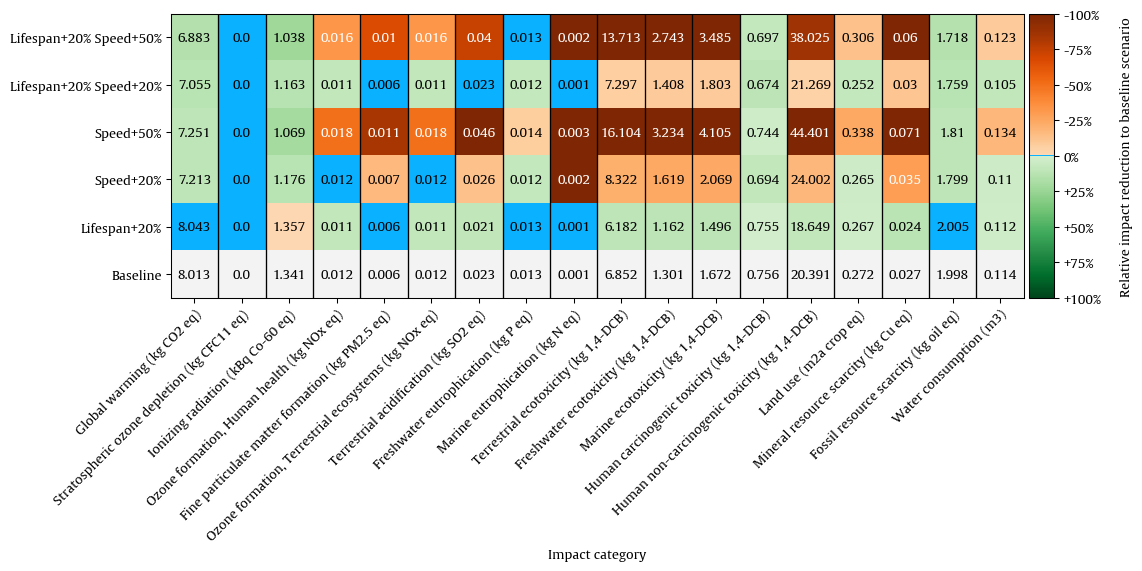}
	\caption{Comparative results for the machining tools case between the baseline and the five DL-based CV-enabled improvement scenarios, for the 18 ReCiPe midpoint indicators.}
	\label{MachinigAllIndicators}
\end{sidewaysfigure}

The baseline scenario has a carbon footprint of 8.013 kg CO\textsubscript{2} eq. to manufacture 100 unit shafts, as described in the FU. Enabled by process improvement with the DL-based CV model for wear assessment, the combination of keeping the cutting tool in use closer to its maximum lifespan (original +20\%) and increasing the process speed by 50\% allows a reduction in the global warming potential of 1.21 kg CO\textsubscript{2} eq. per 100 unit shafts, as illustrated in \Cref{MachinigGlobalWarming}. Considering complementary environmental indicators (see \Cref{MachinigAllIndicators}), the lifespan +20\% and speed +20\% improvement scenario leads to the most significant mitigation of environmental damage. The reduction in the carbon footprint (0.958 kg CO\textsubscript{2} eq., that is around 12\% of the baseline scenario) is slightly lower than in the lifespan +20\% and speed +50\% scenario. However, there is considerably less transfer to other impact categories. Note that only increasing the lifespan of the cutting tool by 20 \%, with the support of the DL-based CV model for wear assessment, is not a relevant strategy in terms of the carbon footprint because of the surplus of impact allocated to the training of the CV model. Consequently, a dedicated LCA is necessary to ensure the environmental benefits of other machining process improvements.

A limitation of this LCA is that we quantify the impact of the cutting tool based on material data from the Ecoinvent database and a literature value \citep{Furberg2019} for hard metal sintering of 11 kWh/kg. However, the impact of an individual tool can vary significantly (\textgreater100\%) depending on the raw materials, production technologies, and energy sources (Alpha).  

\subsubsection{Life Cycle Assessment: Rotating Anodes}
\label{subsubsec:LCA_XRT}
\begin{figure}[htbp]
	\includegraphics[width=\linewidth]{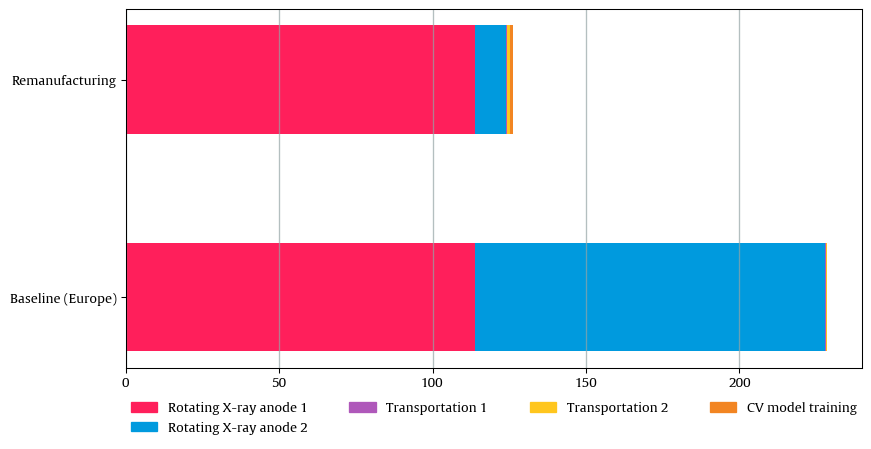}
	\caption{Carbon footprint in the rotating anode case for the baseline and remanufacturing scenario for the European market.}
	\label{XRTGlobalWarmingEuropean}
\end{figure}

\begin{figure}[htbp]
	\includegraphics[width=\linewidth]{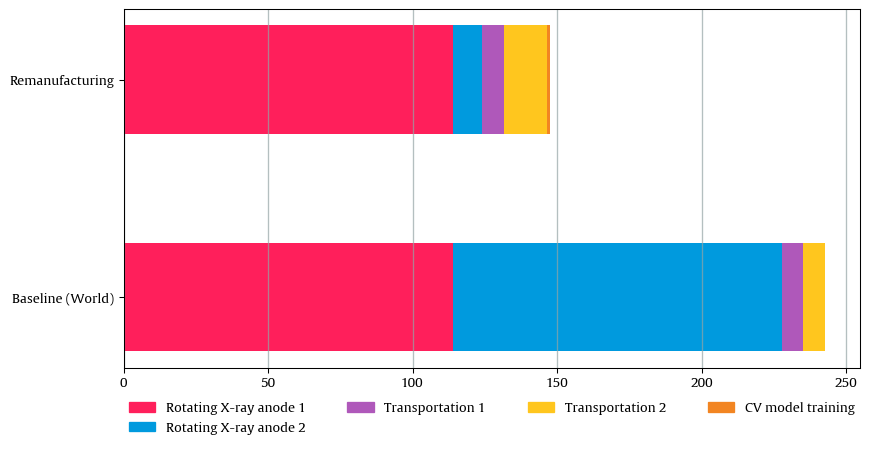}
	\caption{Carbon footprint in the rotating anode case for the baseline and remanufacturing scenario for the non-European market.}
	\label{XRTGlobalWarmingNonEuropean}
\end{figure}

\begin{sidewaysfigure}[htbp]
	\includegraphics[width=\linewidth]{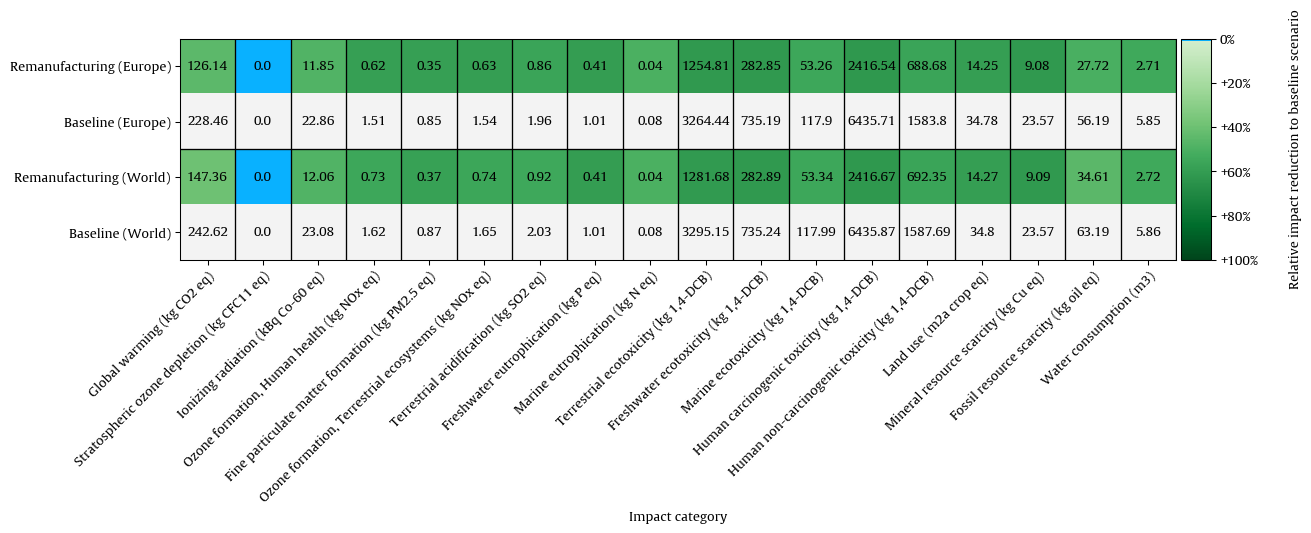}
	\caption{Comparative results for the rotating anode case for the European and non-European market between the baseline and the respective two DL-based CV-enabled improvement scenarios, for the 18 ReCiPe midpoint indicators.}
	\label{XRTAllIndicators}
\end{sidewaysfigure}

In the rotating anode case, the DL-based CV-supported remanufacturing scenario leads to significantly increased environmental sustainability (compare \Cref{XRTGlobalWarmingEuropean} and \Cref{XRTGlobalWarmingNonEuropean} for European and non-European customers, respectively). This is possible because many energy- and resource-intensive processes for producing rotating anodes do not have to be repeated. The carbon footprint is reduced by 44.79 \% / 39.26 \% in the remanufacturing scenario (European/non-European customers). \Cref{XRTAllIndicators} illustrates that no impact transfers occur for the improvement scenario for the rotating anode case.  

\section{Discussion}
\label{sec:Discussion}
In this section, we interpret the obtained results, relate them to similar literature, conceptualize our approach, and explore possible implications. It is our hope that other researchers and practitioners can transfer it to similar scenarios. 

\subsection{Interpretation of Results}
\label{InterpretationResults}
Regarding the CV results, it is essential to note that the detection performance of DL-based CV models will further improve with a higher amount of training data \citep{sun2017revisiting}. 
In terms of the LCAs, several considerations should be remembered. Generally, we want to highlight again that the sustainability improvements cannot be attributed only to the DL-based CV models. To a certain extent, the described improvement scenarios are also possible with a selective manual visual inspection of the wear states of the products. For example, machining processes of customers are already being improved by application engineers. However, the DL-based CV model for wear assessment facilitates further process improvements owing to the efficiency of the CV system. Additionally, domain experts confirm that more effective machining process improvements are possible because of the large number of images that can be assessed efficiently. Consequently, decisions are made on a better statistical basis. Similarly, in the rotating anode case, it is possible to realize the described improvement scenario without a DL-based CV model for wear assessment. However, for this type of product, manual visual inspection would be highly inefficient owing to the image size and the amount of wear to be detected. Consequently, the improvement scenario for the rotating anode case is only partially implemented in the respective industry.

The LCA results indicate a clear improvement in terms of environmental sustainability in the rotating anode case. The remanufacturing of the focal track leads to significant environmental savings (up to 44.79\%) because many resource- and energy-intensive processes necessary to produce rotating anodes do not have to be repeated. 

In comparison, improvements in terms of environmental sustainability in the machining case study seem minor. However, considering the improvements over a more extended period of time, they become considerable. The reduction of almost 1 kg of CO\textsubscript{2} equivalent in the improvement scenario with a 20\% increase in both lifespan and process speed refers to 100 unit shafts. These can be produced in a single hour of production on a single machining center under the baseline scenario. Scaling this to the production of a year (assuming one shift of eight hours and five working days) leads to a saving of 1,993 kg CO\textsubscript{2} equivalent for one machining center. Additionally, machining is an extremely widespread manufacturing process; consequently, these savings can be scaled up to numerous production sites and processes.

For both LCAs, it is interesting to consider the current trend towards an emission-free electricity mix in many countries \citep{IEA2021}. For the rotating anode case, the emissions resulting from transportation will become more relevant. For the machining tools case, the lifespan of the tool will play an even more significant role than it currently does.
\subsection{Related Work: Similarities and Differences}
\label{RelationSimilarLiterature}
Our approach is a type of sustainable smart PSS described by \citet{Li2021}. As proposed by them, we do not focus only on the sustainability of physical materials and components but also consider the information value that can be extracted from physical products.
However, in contrast to previous studies (e.g., \cite{Li2021, zhang2017framework}) and the definition by \citet{Alcayaga2019}, our approach does not rely on smart products in the sense of connected products. Still, we consider our approach as smart because the DL-based CV model allows us to extract relevant information from the considered products in an automated way. Consequently, we extend \citet{Li2021}'s conceptualization to a wider range of products. There is a multitude of reasons why a product is not equipped with sensors or radio frequency identification tags: it might not be economically viable, technically not possible, or not desirable from a data privacy point of view. One might also argue that the camera capturing the pictures is our sensor \citep{martin2021virtual}. Then one could consider our approach as a type of smart retrofitting, which \citet[p. 1]{jaspert2021smart} describe as ``a sustainable approach of transforming the current state of legacy equipment into smart and connected assets.'' Our approach differs from previously documented ones because a static image provides information about the usage stage. In that sense, our analysis is forensic, as we do not have live data about the usage stage or several points in time.
\subsection{Conceptualization of Our Approach}
\label{ConceptualizationApproach}
We believe DL-based CV models can facilitate many more PSSs with reduced environmental impact, for example, by enabling the remanufacturing of products \citep{kjaer2016challenges}. Therefore, in the following section, we describe the prerequisites that need to be fulfilled to make the proposed approach possible.
First, it must be reasonable to implement potential improvements in environmental sustainability, such as re-design, remanufacturing, reuse, and recycling (4R), or a result-oriented PSS based on wear assessment by CV. Hence, the product under consideration must be produced and used in sufficient quantity now and in the future. 
Second, it must be possible to obtain images wherein the wear state of the product can be visually assessed. In this regard, it is also necessary to have sufficient information about the product or usage process to counterbalance the real-life variance in terms of usage and observable wear. For the case of machining tools, this is given because we analyze images of many tools from the same production process. For the rotating anodes, this is given by the large image size. In addition to these technical aspects, organizational aspects are also important. The change from a linear business model to a more circular one requires a willingness to transform of both parties involved ---customer and provider \citep{ceschin2013critical}. Potential economic benefits for both involved companies can incentivize this \citep{fargnoli2018product}. Additionally, co-creation between the provider of the product and the user is helpful, as described by \citet{arnold2017fostering}. In our cases, the providing company can perform the wear state assessment and draw relevant insights based on their domain knowledge. However, most business decisions benefit from additional metadata such as usage process parameters, which typically belong to the company using the product.\\
As illustrated by the results of the LCA for the machining tools case, and described by e.g., \citet{kjaer2016challenges} and \citet {Tukker2015} PSSs have the potential to yield improvements of environmental benefits but of course cannot guarantee these improvements. Hence, it is essential to perform dedicated LCAs. According to \citet[p. 95]{kjaer2016challenges} three relevant scopes are distinguished, where LCAs may be applied to evaluate PSSs: ``(1) evaluating options within the PSS itself; (2) comparing a PSS with an alternative; and (3) modelling the actual contextual changes caused by the PSS.''

\subsection{Implications}
\label{Implications}
As demonstrated in this work, DL-based CV models can facilitate sustainable smart PSSs. This can yield numerous benefits for the manufacturing industry. Primarily, providers and consumers can reduce the environmental impact of their respective overall systems. Additionally, economic benefits can be expected. Domain experts confirm for both case studies that economic benefits are anticipated for providers and customers. Additionally, existing literature \citep{Annarelli2016, yang2019product} confirms that result-oriented PSSs can yield economic advantages. As outlined before, DL-based CV models can be used to address uncertainties regarding risks and costs that often hinder the formation of result-oriented PSSs \citep{Erkoyuncu2011}.

\section{Conclusion}
\label{sec:Conclusion}
In this work, we demonstrate the effectiveness of deep-learning-based computer vision, a special type of artificial intelligence, for facilitating sustainable smart product-service systems. To this end, we perform two case studies: one on machining tools and another on rotating X-ray anodes. For both case studies, we first demonstrate the feasibility of detecting the wear state with deep-learning-based computer vision as an input for sustainable smart product-service systems. Subsequently, we perform life cycle assessments based on real data and the inputs of domain experts. The results demonstrate the possible improvements in environmental sustainability resulting from sustainable smart product-service systems based on deep learning-based computer vision.

A limitation of our work is its focus on the environmental dimension of sustainability. The concept of sustainability typically consists of three pillars: economic, environmental, and social. Economic sustainability was not evaluated explicitly in this work; however, as described previously, benefits are expected according to domain experts and the existing literature. Although we did not explicitly evaluate social sustainability, a direct effect on this dimension is not expected. Particularly, the proposed approach aims not to fully automate jobs but to complement human experts in tedious jobs and free their capacity for jobs that are more suited to their skill levels. An additional limitation is that the validity of the case study-based research at hand is limited to the cases' contexts. 

We hope that this work will inspire researchers and practitioners to conduct similar studies and look forward to studies extending sustainable smart product-service systems to products that are not inherently smart. This can help the manufacturing industry reduce its environmental impact while increasing its competitiveness. More broadly, we hope that this work accelerates the implementation of novel ideas and artificial-intelligence-based innovations that have a positive environmental impact. 

\section*{Declaration of competing interest} The authors declare that they have no known competing financial interests or personal relationships that could have appeared to influence the work reported in this paper.

\section*{Acknowledgements}
We thank Ceratizit Austria GmbH (in particular Jonathan Schäfer) and Plansee SE (in particular Jasmin Matti) for facilitating and supporting this research.

\section*{Appendix}
\subsection*{Energy consumption for training of CV models}
\label{EnergyConsumptionCV}
In our improvement scenarios, there are several computer-supported phases: 
Training of the DL-based CV model, prediction of the DL-based CV model, and execution of the user-centric artifact. The energy consumption of the latter two is negligible because these applications can run on a standard laptop in parallel to other tasks. Training of the DL-based CV model is the most energy-intensive step because it must be performed on specialized hardware. For both machining tools and rotating anodes, approximately five training runs are required to find a suitable model. For the machining tools case, a training run takes 25 minutes, and for the rotating anodes, it takes 30 minutes. The IBM AC922 machine with two graphical processing units we used consumes a maximum of 1.15\,kW.
Multiplied by five training runs and 25/30 minutes per training, we obtain 2.395\,kWh and 2.875\,kWh for the machining tools and the rotating anode cases, respectively.

\bibliographystyle{cas-model2-names}

\bibliography{AIForSustainability}

\end{document}